\documentclass[lettersize,journal]{IEEEtran}
\usepackage{amsmath,amsfonts}
\usepackage{array}
\usepackage[caption=false,font=normalsize,labelfont=sf,textfont=sf]{subfig}
\usepackage{textcomp}
\usepackage{stfloats}
\usepackage{url}
\usepackage{verbatim}
\usepackage{graphicx}
\usepackage{cite}
\usepackage{algpseudocode}
\usepackage[table]{xcolor}
\usepackage{cite}

\usepackage{makecell}  
\usepackage{setspace}  
\setcellgapes{5pt}
\usepackage{amsmath,amsfonts}
\usepackage{array}
\usepackage[caption=false,font=normalsize,labelfont=sf,textfont=sf]{subfig}
\usepackage{textcomp}
\usepackage{stfloats}
\usepackage{url}
\usepackage{verbatim}
\usepackage{graphicx}

\usepackage{multirow}
\usepackage{multicol}
\usepackage{float}
\usepackage{array}
\usepackage{booktabs}
\usepackage[utf8]{inputenc}
\usepackage[ruled,linesnumbered]{algorithm2e}
\usepackage{makecell}
\usepackage {booktabs}
\usepackage{url}
\usepackage{graphicx}
\usepackage{algpseudocode}
\usepackage{float}
\usepackage[utf8]{inputenc}
\usepackage{soul, color, xcolor}
\usepackage{xcolor}
\usepackage{colortbl}
\usepackage[table]{xcolor}
\usepackage{bbding}
\usepackage{pifont}
\usepackage{arydshln} 
\usepackage{xcolor}
\usepackage{soul}

\usepackage{booktabs}

\newcommand{\lkj}{\color{blue}}

\def\BibTeX{{\rm B\kern-.05em{\sc i\kern-.025em b}\kern-.08em
    T\kern-.1667em\lower.7ex\hbox{E}\kern-.125emX}}
\usepackage{balance}

\begin{document}


\title{Smile on the Face, Sadness in the Eyes: Bridging the Emotion Gap with a Multimodal Dataset of Eye and Facial Behaviors}

\author{Kejun Liu, Yuanyuan Liu*,~\IEEEmembership{Member,~IEEE,}, Lin Wei, Chang~Tang,~\IEEEmembership{Senior Member,~IEEE,} Yibing Zhan,~\IEEEmembership{Member,~IEEE,} Zijing Chen,~\IEEEmembership{Member,~IEEE,} Zhe Chen, ~\IEEEmembership{Member,~IEEE,} 

  \vspace{-1em}
 \IEEEcompsocitemizethanks{
  \IEEEcompsocthanksitem Kejun Liu, Yuanyuan Liu, and Lin Wei are with the School of Computer Science, China University of Geosciences (Wuhan), China. E-mail: {liukejun, liuyy, linw}@cug.edu.cn.
  \IEEEcompsocthanksitem Chang Tang is with the School of Software Engineering, Huazhong University of Science and Technology, China. E-mail: tangchang@cug.edu.cn.
 \IEEEcompsocthanksitem Yibing Zhan is with the School of Computer Science, Wuhan University, China. E-mail: zybjy@mail.ustc.edu.cn.
\IEEEcompsocthanksitem Zijing Chen and Zhe Chen are 
with the School of Computing, Engineering and Mathematical Sciences, La Trobe University, Australia. They are also with the Cisco-La Trobe Centre for Artificial Intelligence and Internet of Things. E-mail: {zijing.chen, zhe.chen}@latrobe.edu.au.
}

\thanks{*Corresponding author: Yuanyuan Liu}
}

\markboth{Journal of \LaTeX\ Class Files,~Vol.~14, No.~8, August~2023}%
{Shell \MakeLowercase{\textit{et al.}}: A Sample Article Using IEEEtran.cls for IEEE Journals}

\maketitle

\begin{abstract}
Emotion Recognition (ER) is the process of analyzing and identifying human emotions from sensing data. Currently, the field heavily relies on facial expression recognition (FER) because visual channel conveys  rich emotional cues. However, facial expressions are often used as social tools rather than manifestations of genuine inner emotions. To understand and bridge this gap between FER and ER, we introduce eye behaviors as an important emotional cue and construct an Eye-behavior-aided Multimodal Emotion Recognition (EMER) dataset. 
To collect data with genuine emotions,  spontaneous emotion induction paradigm is exploited with stimulus material, during which non-invasive eye behavior data, like eye movement sequences and eye fixation maps, is captured together with facial expression videos.  To better illustrate the gap between ER and FER, multi-view emotion labels for mutimodal ER and FER are separately annotated. 
Furthermore, based on the new dataset, we  design a simple yet effective Eye-behavior-aided MER Transformer (EMERT) that enhances ER by bridging the emotion gap. EMERT leverages modality-adversarial feature decoupling and a multitask Transformer to model eye behaviors as a strong complement to facial expressions.  
In the experiment, we introduce seven multimodal benchmark protocols for a variety of comprehensive evaluations of the EMER dataset. The results show that the EMERT outperforms other state-of-the-art multimodal methods by a great margin, revealing the importance of modeling eye behaviors for robust ER.  
To sum up, we provide a comprehensive analysis of the importance of eye behaviors in ER, advancing the study on addressing the gap between FER and ER for more robust ER performance. 
Our EMER dataset and the trained EMERT models will be publicly available at https://anonymous.4open.science/r/EMER-database.

\end{abstract}


\vspace{-2em}
\section{Introduction}
\IEEEPARstart{E}{motion} recognition (ER) aims to understand and identify human psycho-emotional states across diverse behaviors and contexts, playing a key role in human-computer interaction and cognitive science \cite{10812840}. 
It also finds broad applications in multimedia scenarios and applications, such as video surveillance, intelligent education systems, affective healthcare, and personalized advertising \cite{MMA-DFER}.
Recent advances in Facial Expression Recognition (FER) have driven progress in ER \cite{10404024,10814686,liu2023expression}, as facial expressions are widely regarded as strong indicators of emotional states. Most FER-based ER approaches rely on visual cues from images or videos \cite{NGAI2022107,10552397}, supported by both static and dynamic FER datasets. Static datasets like SFEW \cite{dhall15} and JAFFE \cite{lyons98} consist of still facial images with emotion labels, while dynamic datasets such as AFEW 7.0 \cite{dhall17} and DFEW \cite{jiang20} provide temporally evolving facial expressions from videos.
Using these datasets, many FER-based ER methods have been developed to classify Ekman’s six basic emotions (\textit{i.e.}, happiness, sadness, fear, surprise, disgust, and anger) \cite{ekman93}. However, relying solely on visual facial expressions may be insufficient, as such cues can be consciously masked or suppressed, leading to unreliable recognition in certain contexts \cite{NGAI2022107}.

\begin{figure}[t]
  \centering
   \includegraphics[width=1.0\linewidth]{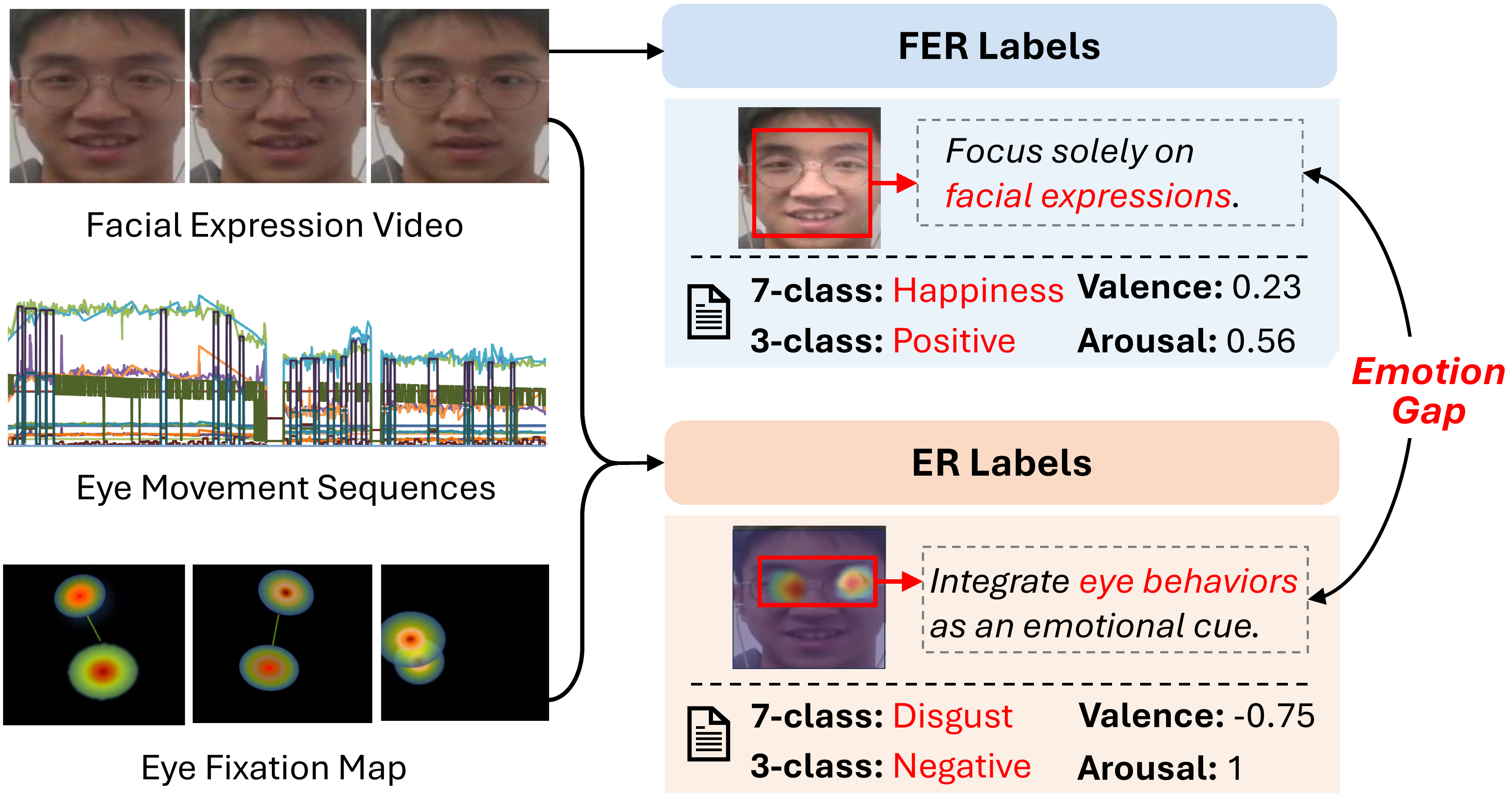}
   \vspace{-2em}
  \caption{An example from our EMER dataset.  EMER comprises  facial expression videos, eye movement sequences, and  eye fixation maps, along with multi-view emotion annotations, including FER labels and ER labels, providing more comprehensive emotion analysis.}

  \label{fig:example}
  \vspace{-2em}
\end{figure}

In the area of ER, existing literature has shown that relying solely on visual facial expression signals is inadequate due to the subjective and camouflaged nature of facial expressions \cite{10812840,10491325}. This limitation results in a significant `\textit{\textbf{emotion gap}}' when applied to various scenarios.
Here, the \textit{\textbf{emotion gap}} refers to the disparity between facial expressions and genuine emotions of individuals \cite{na}. 
As an intuitive example, when a person is concealing his or her sad feelings, he or she may put a big smile on the face as a natural response.  In such scenarios, conventional FER methods would be misled by the smile expression and could not robustly recognize the true sadness emotion. To bridge this \textit{\textbf{emotion gap}} between FER and ER, 
recent studies have explored the integration of additional physiological modalities, such as electroencephalogram (EEG) signals and eye behavior signals \cite{5975141,5871728,8283814}, which are more intuitively reflective of the true state of human emotions. These efforts have led to more robust and comprehensive multimodal ER (MER) systems.
To achieve this, some physiological signals enhanced  MER datasets have been developed, as shown in Table ~\ref{tab:existing_databases}.  
For example, the DEAP \cite{5871728}  collects facial expression videos and EEG signals from 32 participants, 
and MAHNOB-HCI dataset \cite{5975141}  contains eye movement data and  facial images from 27 participants.
Although these MER datasets help mitigate the gap between expression and emotion, most they rely on specialized and expensive sensing devices to capture high-quality EEG or eye signals. This makes these \textit{\textbf{existing datasets collected on a small scale, with a relatively homogenous set of participants and labeling}}. This limits the scalability and utility of such systems for real-world applications. 


\begin{table*}
\vspace{-2em}
	\small
	\caption{Summary of existing popular multimodal emotion datasets and our proposed EMER dataset.}
    \vspace{-1em}
	\label{tab:existing_databases}
        \resizebox{\linewidth}{!}{
	\begin{tabular}{c|c|c|c|c|c|c|c|c}
		\hline
		\multirow{1}*{\textbf{Dataset}} & \multirow{1}*{\textbf{\#Part.}}  & \multirow{1}*{\textbf{Data Quality}} & \multirow{1}*{\textbf{Non-invasive Sensor}} & \multirow{1}*{\textbf{Visual Facial Images}} & 
        \multirow{1}*{\textbf{Eye Movement Data}} 
        & \multirow{1}*{\textbf{Eye Fixation Maps}} 
        & \multirow{1}*{\textbf{Both ER \& FER Anno.}} &
        \multirow{1}*{\textbf{Emotion Gap Analysis}} \\

        \hline

        \multirow{1}*{CMU-MOSI \cite{vasileios_skaramagkas_2022_7794625}} &N/A  & Noisy & \checkmark & \checkmark &   & & & 
        \\ 

        \multirow{1}*{CMU-MOSEI \cite{zadeh2018multimodal}} &N/A & Noisy & \checkmark & \checkmark &    &  & & 
        \\ 

        \multirow{1}*{IEMOCAP \cite{zheng2015investigating}} & 10 & Clean & \checkmark & \checkmark &   & &  &  
        \\ 

        \multirow{1}*{eNTERFACE’05 \cite{1623803}} & 42& Clean & \checkmark  & \checkmark &   &  & & 
        \\ 

        \multirow{1}*{DECAF \cite{7010926}} & 30 & Clean &  & \checkmark & &  &  &
        \\ 

        \multirow{1}*{DEAP \cite{liu2021comparing}} & 32  & Clean &  & \checkmark & &   & &
        \\ 

        \multirow{1}*{SEED-IV \cite{8283814}} & 15 & Clean &   &   &\checkmark&&    &
        \\ 
 
        \multirow{1}*{SEED-V \cite{SEED_V}} & 20 & Clean &   &  &\checkmark &&   &
        \\ 

        \multirow{1}*{MAHNOB-HCI \cite{5975141}} & 27& Clean &     &\checkmark &\checkmark & & & 
        \\ 
        \rowcolor{green!25}
        {\textbf{Our EMER}} & \textbf{121} & Clean & \checkmark & \checkmark & \checkmark & \checkmark &\checkmark& \checkmark
        \\ 
		\hline
		
	\end{tabular}
 }

 \vspace{-1.5em}
\end{table*}

To address the limitations, in this paper, we construct an Eye-behavior-aided Multimodal Emotion Recognition dataset (\textbf{EMER}) featuring larger scale, diverse participants, and multi-view annotations. EMER captures rich emotional cues by integrating facial expression videos, eye movement sequences, and eye fixation maps, along with multi-view emotion annotations, including FER labels and ER labels.  The inclusion of such eye behaviors is inspired by  Hess \textit{et al.} \cite{Hess1960PupilSA} and other psychological studies \cite{10224292}, which demonstrate that eye movements and fixation patterns serve as natural, intuitive responses to emotional states.

To construct the EMER dataset, we adopt a stimulus-induced spontaneous emotion elicitation protocol. Four emotion experts first curated 28 emotional video clips, which were then shown to 121 participants to induce short-term, spontaneous emotional responses. This process yielded 1,303 high-quality multimodal sequences, simultaneously capturing eye behaviors and facial expressions using a non-invasive Tobii Pro Fusion eye tracker\footnote{\url{https://www.tobii.com/products/eye-trackers/screen-based/tobii-pro-fusion}\label{web}} and a high-definition camera. 
Each sample in EMER is annotated with both emotion and facial expression labels using a combined labeling strategy to ensure accuracy and depth, enabling a detailed analysis of the \textit{\textbf{emotion gap}} between FER and ER. To our knowledge, EMER is the first eye-behavior-aided  multimodal dataset specifically designed for both ER and FER, offering unique insights into this gap. Fig.\ref{fig:example} illustrates the multi-view annotations in EMER, and Table ~\ref{tab:existing_databases} compares EMER with existing MER datasets.

In addition, based on the new EMER dataset, we design a simple yet effective  Eye-behavior-aided MER Transformer (\textbf{EMERT})  method. 
Rather than the existing multimodal methods \cite{10109845,10814059},  our EMERT  applies adversarial learning and multi-task Transformer to help explicitly extract modality-complementary affective features, so that the gap between facial expression information and eye behavior information can be better modeled and bridged for more effective ER, providing a strong benchmark for future research.   

To sum up, we summarize the key contributions as follows:
\begin{itemize}
    \item We create EMER, a novel eye-behavior-aided multimodal ER dataset containing 1,303 spontaneous samples from 121 participants. EMER includes eye movement sequences, eye fixation maps, and facial expression videos, with both FER and ER labels to enable comprehensive emotion gap analysis. To our knowledge, EMER is the first dataset of its kind, offering a new direction for emotion gap research in ER. 
    \item The EMER dataset introduces comprehensive annotation strategies for achieving multi-view emotion labels. We cover 3-class coarse ER and FER labels (namely positive, negative and neutral), 7-class fine ER and FER labels (namely happiness, sadness, fear, surprise, disgust, anger, and neutral), 2-dimensional continuous emotion ratings (valence and arousal), as well as facial expression intensity (0-3). All of the annotation information contributes to the explicit investigation of the emotion gap, aiming to delve into the details of how to improve ER with multimodal data. 
    \item We devise a simple yet effective benchmarking method, EMERT, to achieve robust ER performance by explicitly and effectively bridging the emotion gap between facial expressions and eye behaviors. The EMERT has shown significant benefits in ER. 
     \item We carried out a comprehensive evaluation of various multimodal methods on our EMER dataset, with seven benchmarking protocols. By addressing the gap between FER and ER, we can further demonstrate that both two tasks benefits from the emotional cues from multimodal data, highlighting the importance of explicitly analyzing the emotion gap for future research.  
   
\end{itemize}

\vspace{-1em}
\section{Related Work}

\subsection{\textbf{Facial Expression-based Multimodal Emotion Datasets}}

Currently, there are two main types of facial expression-based multimodal emotion datasets, namely in-the-wild collected emotion datasets \cite{vasileios_skaramagkas_2022_7794625} and lab-induced
spontaneous emotion datasets \cite{5975141,liu2021comparing}. 
In-the-wild collected emotion datasets mainly contain facial expression data, audio, and text gathered from the web or social medias. These datasets often contain various sources of noise and can be challenging to annotate accurately, ultimately compromising their utility for complex applications in emotion recognition.
For example, CMU-MOSI \cite{vasileios_skaramagkas_2022_7794625} consists of 2199 clips with video, audio, and text data, collected from YouTube and annotated with emotional scores in the range [-3,3]. 
Lab-induced spontaneous emotion datasets contain facial expressions and other physiological signals, such as EEG, Electrocardiography(ECG), Galvanic Skin Response(GSR), and so on. 
DEAP \cite{liu2021comparing} contains 1280 multimodal samples from 32 participants, with annotations for valence, arousal, dominance, likability, and familiarity, along with facial videos and physiological signals (EEG, GSR, ECG). 
MAHNOB-HCI \cite{5975141} includes 565 samples from 27 participants, each with eye movements, EEG, physiological signals, video, audio, and labels for 9-class emotion, valence, arousal, dominance, and predictability.
Despite progress, the former suffers from web-induced noise, while the latter is constrained by limited scale and participant diversity.

\vspace{-1em}
\subsection{\textbf{Facial Expression Recognition}}

The FER task aims to understand individual emotions from his/her visual facial expressions. 
Currently methods are divided into two main categories, namely static FER methods and dynamic FER methods \cite{9157210, liu2023expression}.
Static FER methods focus on recognizing facial expressions from static face images and have achieved significant achievements. 
Wang \textit{et al.} \cite{9157210} recognized facial expressions from low-quality images by introducing an effective self-healing network (SCN). 
By Contrary, dynamic FER methods explores spatio-temporal information from video sequences, obtaining more robust FER performance. 
Ma \textit{et al.} \cite{unknown} proposed the spatial-temporal Transformer to capture discriminative emotion features within each frame and model contextual relationships among frames. 
Liu \textit{et al.} \cite{liu2023expression} proposed Expression snippet Transformer (EST) to decompose videos into expression snippets to enhance intra- and inter-snippet visual modeling capabilities, respectively.
Although progress has been made in FER, these methods can only recognise information from facial expressions, which can be easily camouflaged in some scenes, and this can lead to recognition results that deviate from the true emotion.

\vspace{-1em}
\subsection{\textbf{Multimodal Emotion Recognition}}
 
Multimodal ER aims to predict human emotions from multiple modalities, such as video, audio, and physiological signals.
Most existing methods mainly are divided into two categories, \textit{i.e}, representation learning-based methods \cite{Representation,DBLP:conf/aaai/ZhaoMGYXXHCK20} and multimodal fusion-based methods \cite{XIE2023126649}. Representation learning-based methods focus on learning specific modality representations by considering the difference and consistency of different modalities, thus improving multimodal emotion recognition. For example, VAANET~\cite{DBLP:conf/aaai/ZhaoMGYXXHCK20}, which integrated spatial, channel-wise, and temporal attentions for audio-video emotion recognition. Multimodal fusion-based methods attempt to learn the interactive information between different modalities by designing complex fusion mechanisms. For example, MulT~\cite{MulT} used a set of Transformer encoders to capture both unimodal and cross-modal interactions. Kernel-based Extreme Learning Machine (ELM) \cite{duan2016multimodal} recognized video emotions by combining video content and EEG signals.
Despite the progress, most methods did not consider the raw noises existing in the specific modality features, leading to sub-optimal results.

\begin{figure*}[ht]
  \centering
  \vspace{-1em}
  \includegraphics[width=1.0\linewidth]{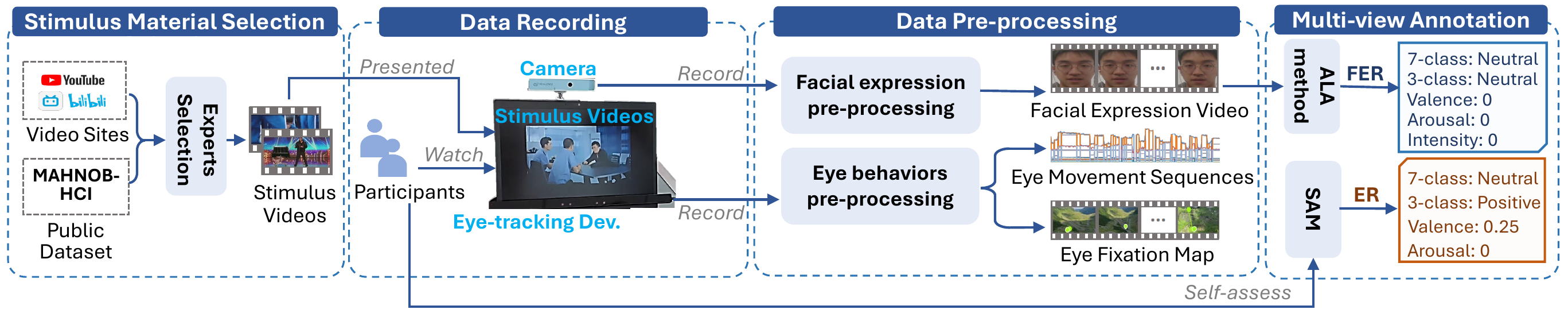}
  \vspace{-2em}
  \caption{The collection framework for our EMER dataset. The EMER dataset is  multimodal, participant-rich, and multi-view annotation emotion dateset, providing a novel research direction in understanding the emotion gap between ER and FER. }
  \label{fig: model_structure}
  \vspace{-1em}
\end{figure*}



\vspace{-1em}
\section{Proposed EMER Dataset}
To gain deeper insights into the gap between FER and ER, we construct the Eye-behavior-aided Multimodal Emotion Recognition (EMER) dataset. As illustrated in Fig.~\ref{fig: model_structure}, the construction pipeline comprises four stages: stimulus selection, data recording, data pre-processing, and multi-view annotation. Through this pipeline, we collects 1,303 spontaneous emotional sequences collected from 121 participants, covering 3 modalities: facial expression videos, eye movement sequences, and eye fixation maps. In addition, we provide both ER and FER labels using distinct annotation strategies, enabling a comprehensive analysis of the gap between FER and ER.


\vspace{-1em}

\subsection{\textbf{Stimulus Material Selection and Participants}}
Following the protocols of MAHNOB-HCI \cite{5975141} and SEED \cite{zheng2015investigating}, we adopt a stimulus material-induced paradigm to elicit spontaneous emotions. We first collect 115 candidate videos across seven basic emotion categories from public datasets and platforms such as BiliBili and YouTube. Four emotion experts then select the top 4 most emotionally evocative clips per category, yielding 28 final stimuli (1–2 minutes each). 
Some examples are shown in Fig.\ref{fig:stimulus}., with details in the \textit{supplementary material (Sec.I)}. 
We recruit 121 participants (76 males, 45 females; aged 18–40) from diverse backgrounds. In a controlled lab environment, each participant views the selected stimuli to evoke short-term, spontaneous emotional responses. All participants provided informed consent in accordance with GDPR\footnote{\url{https://gdpr-info.eu/}}, with the consent form included in the supplementary material.


\begin{figure}[h]
  \centering
  \includegraphics[width=0.7\linewidth]{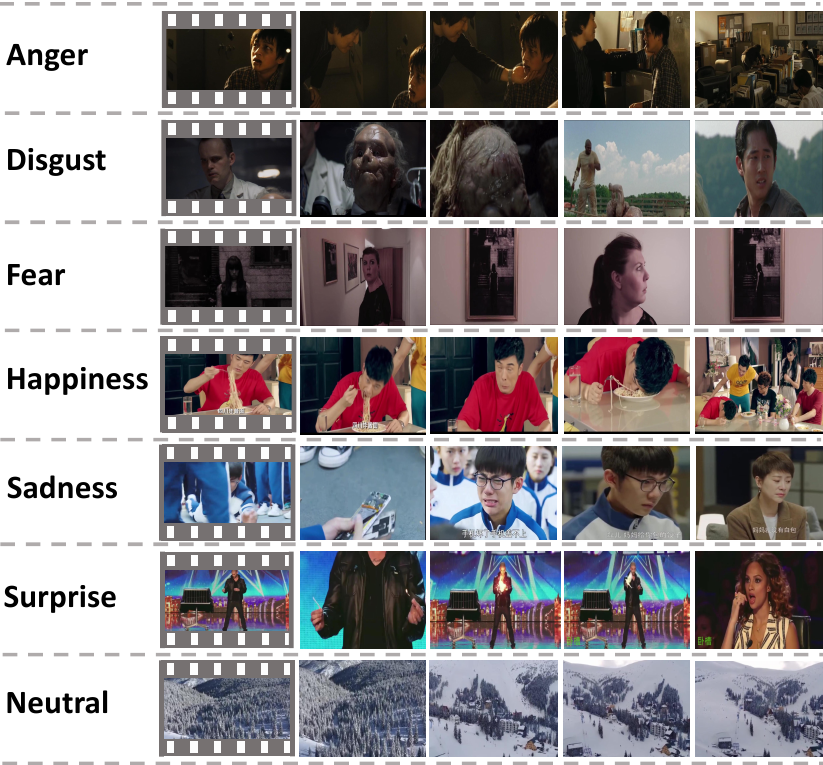}
  \vspace{-1em}
  \caption{ Some examples of stimulus materials.}
  \label{fig:stimulus}
  \vspace{-1em}
\end{figure}

\vspace{-1em}
\subsection{\textbf{Data Recording}}
Using the selected emotion-stimulus videos, each participant is asked to watch different categories of stimulus videos in sequence, and his eye movement sequences, emotion-related eye fixation maps, and facial expression videos are recorded by a Tobbi Pro Fusion eye-tracking device 
and a high-definition camera, simultaneously. 
After viewing the stimulus videos, each participant completes an emotional self-assessment questionnaire for each stimulus video. 
Ultimately, we collected 1,623 original multimodal data samples from the 121 participants. {\textit{More details can be seen in our supplementary material (Sec.II).}}


\vspace{-1em}
\subsection{\textbf{Data Pre-processing}}
To maintain the integrity and synchronization of the collected multimodal data, we meticulously align, trim, and filter the original data, obtaining a total of 1,303 high-quality multimodal emotional data samples processed for our EMER.

\subsubsection{\textbf{Facial expression pre-processing}}
Due to various emotion-irrelevant visual noise (\textit{e.g.}, illumination, head poses, etc.) in raw facial data, we carry out a 2-step pre-processing.
In the first step, we use illumination normalization \cite{ying2017new} to remove lighting variations across different frames of a video. Then, we employed a state-of-the-art deep learning model, MTCNN \cite{zhang2016joint}, to extract facial landmarks and we then performed face alignment according to the landmarks to ensure consistency across all video frames.


\subsubsection{\textbf{Eye behaviors pre-processing}}
To align unsynchronised eye behavior signals, we employ a 3-step process to perform blink correction, sweep correction, and pupil correction\footnote{Pupil correction: Pupil data alone may contain more noise, while pupil fluctuation is more capable of expressing emotion\cite{NGAI2022107}, pupil correction is performed to replace pupil data with pupil fluctuation.} and align asynchronous eye movement sequences and eye fixation maps. 
Regarding blink correction, we first identify eye movement sequences with blink durations outside the range of 75 ms and 425 ms as invalid blinks \cite{NGAI2022107}, and then use linear interpolation  to correct the invalid blinks for blink correction. 
In sweep correction, linear interpolation is also used to correct the eye sweep data. Lastly, following \cite {NGAI2022107},  we use the difference between the pupil diameters corresponding to the current and the previous timestamp for pupil correction in eye movement sequences.
It is worth mentioning that, to make the eye fixation maps related to emotions, we remove invalid eye fixations according to invalid eye movements.


\vspace{-1em}
\subsection{\textbf{Multi-view Annotation}}
As introduced earlier, we provide multi-view emotion annotation with both ER and FER to help analyze the gap between emotions and facial expressions. To further clarify, the ER labels are based on participant- and organizer-inducing emotion annotations, while FER labels are assigned by experts through analysis of recorded facial videos. 
Here, we will explain label formats and annotation methods in detail. 

\subsubsection{\textbf{Label Formats.}}
Each ER label contains three key aspects: (1) 3-class coarse ER labels, \textit{i.e.}, positive, negative, and neutral; (2) 7-class fine ER labels, \textit{i.e.}, happiness, sadness, fear, disgust, surprise, anger, and neutral; (3) valence and arousal ratings in the range [-1,1], where a higher valence score signifies a greater level of happiness, while a higher arousal score indicates a greater degree of excitement. Meanwhile, our EMER also offers four FER annotations (see Fig.\ref{fig:example}), which consist of 3-class coarse FER labels, 7-class fine FER labels, valence, and arousal ratings within the [-1, 1] range, and facial expression intensity within the [0,3] range. 
Although ER and FER labels are obtained separately, it is important to note that ER and FER labels come from the shared set of emotional categories, such as happiness, sadness, fear, surprise, disgust, anger, and neutral within  7-class fine labels, as well as positive, negative, and neutral within  3-class coarse labels. 

In the rest of this paper, we use $j$ to index each data sample, and we utilize $e_j$ and $f_j$ to represent the corresponding ER label and FER label, respectively.

\subsubsection{\textbf{Annotation Method}}
To achieve multi-view emotion annotation, we introduce two different annotation strategies to provide ER and FER labels, respectively,  for comprehensive emotion analysis.

\begin{figure}[h]
  \centering
  \vspace{-0.5em}
  \includegraphics[width=0.8\linewidth]{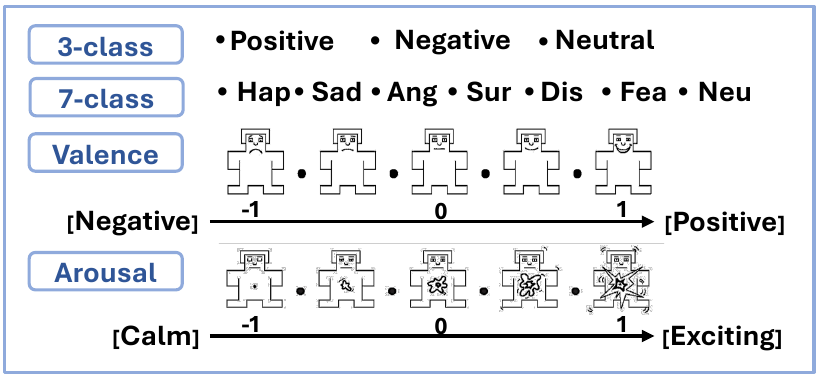}
  \vspace{-1em}
  \caption{The SAM self-assessment for the ER annotation. } 
  \label{fig:sam}
  \vspace{-0.5em}
\end{figure}

\underline{\emph{ER annotation.}} Following the annotation methods in  \cite{5975141}, \cite{zheng2015investigating}, 
we annotate ER labels $e_j$.  Using participant self-assessments via the Self-Assessment Manikin (SAM) \cite{liu2021comparing}, as shown in Fig.\ref{fig:sam}. For each collected data sample,  participants use the SAM to rate their emotional states, ensuring that the ER labels reflect both individual experiences and stimulus effects—providing rich, human-centered annotations for multimodal emotion data.  

\begin{figure}[h]
  \centering
  \includegraphics[width=0.85\linewidth]{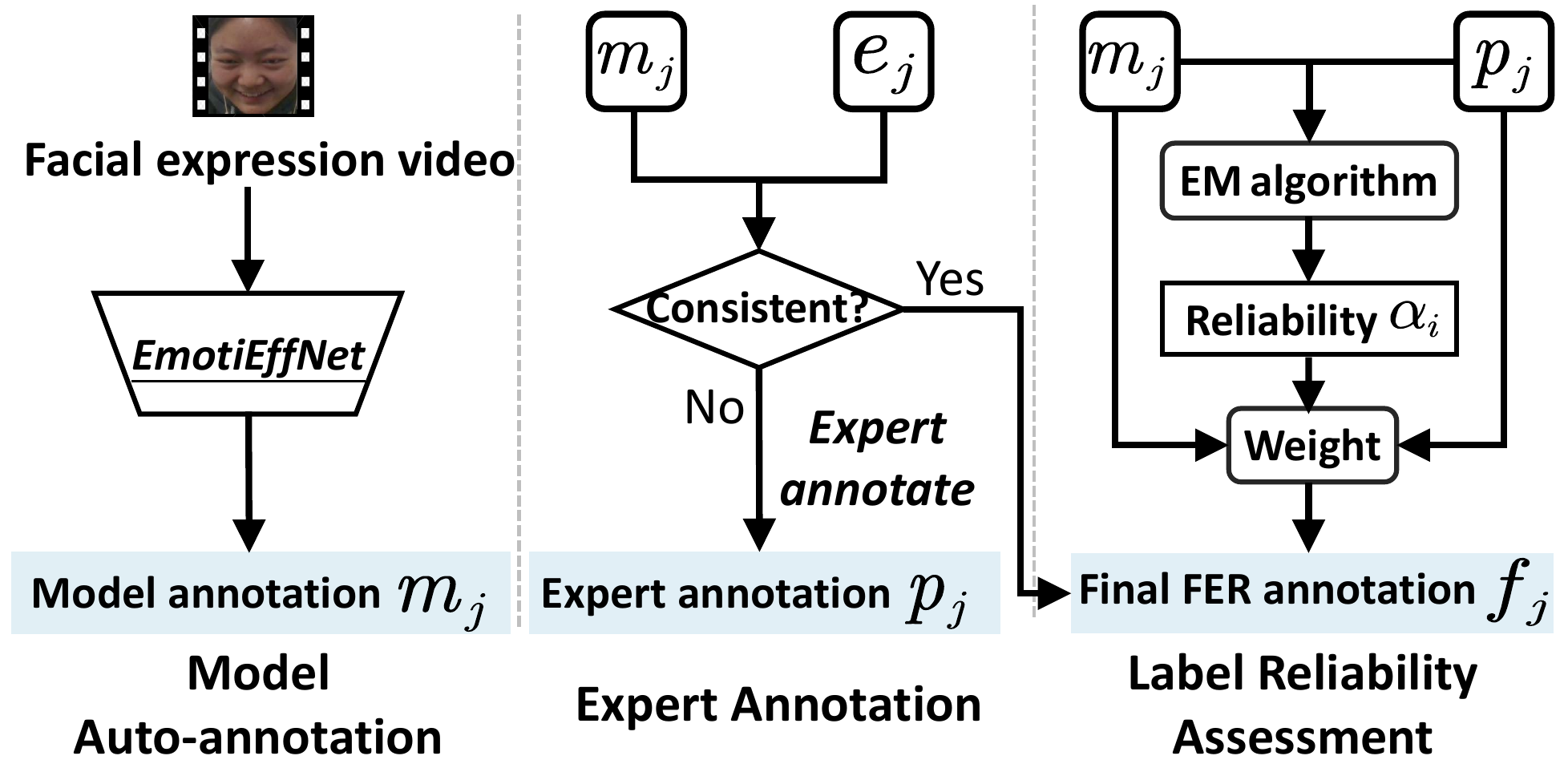}
  \vspace{-1em}
  \caption{The ALA pipeline for high-reliability FER annotation, including 
  model auto-annotation, expert annotation, and annotation reliability assessment. } 
  \label{fig:anotation}
  \vspace{-1em}
\end{figure}

\underline{\emph{FER annotation.}}  Manual FER annotation is often labor-intensive, time-consuming, and prone to subjectivity \cite{zadeh2018multimodal}, \cite{liu2021comparing}. To mitigate  this issue, we adopt an Active Learning-based Annotation method (ALA) that combines deep model auto-annotation with manual expert annotation. As illustrated in Fig.\ref{fig:anotation}, ALA involves model auto-annotation, expert annotation, and label reliability assessment, resulting in efficient and high-quality FER labels.

Model auto-annotation represents the process of using machines to annotate facial videos. We employ EmotiEffNet \cite{savchenko2022classifying}, a pre-trained deep neural network specialized in FER, to automatically annotate each collected data, resulting in the model-generated FER labels denoted $m_j$. This significantly expedites the annotation process, leading to substantial time and resource savings. 

After model auto-annotation, there would be inherent prediction biases of machine models, and we enhance the machine-generated labels by incorporating expert annotations. Specifically, we first use ER labels to identify and filter inconsistent model-generated FER labels, and the remaining FER labels that are consistent with ER labels are directly stored as annotated FER labels. Due to the emotion gap, there could be many inconsistent labels. For these inconsistent labels, we enlist the expertise of four emotion specialists for re-annotation, ultimately yielding the ER label set termed as $T_j = \{m_j, p_j \}$, where $p_j$ represents expert labels and $m_j$ represents model-generated labels consistent with ER labels.

Using the FER label set $T_j$, we then employ the EM algorithm \cite{li2017reliable} to assess annotation reliability, enhancing the quality and reliability of FER labels. The EM algorithm comprises two key optimization steps: the E-step and the M-step. The E-step calculates the posterior probability for potential correct annotations, while the M-step optimizes the log-likelihood of each label by estimating each label reliability ${\alpha }_{i}$ within $T_j$.
By iteratively cycling through E-step and M-step until convergence,  we  ultimately obtain high-quality FER labels $f_j$ through a weighted voting process, which can be formulated as: $f_j = \underset{i=1}{\overset{5}{\mathop \sum }}\,{\frac{{{\alpha }_{i}}}{\mathop{\sum }_{i=1}^{5}{{\alpha }_{i}}} \cdot t_{j}^{i}}$, where $t_{j}^{i} \in T_j$.    

\begin{figure*}[t]
  \centering
\vspace{-1.5em}
  \includegraphics[width=0.98\linewidth]{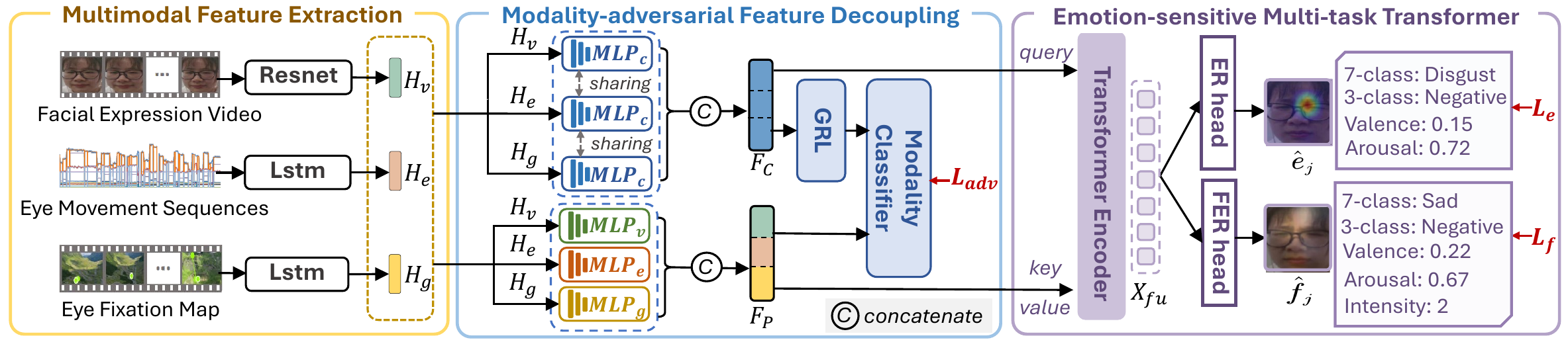}
  \vspace{-1em}
  \caption{The general framework for our EMERT method. The EMERT achieves robust ER performance by explicitly and effectively bridging the emotion gap between facial expressions and eye behaviors. }
  \label{fig: model}
  \vspace{-1.5em}
\end{figure*}


\vspace{-1em}
\subsection{\textbf{Metadata in EMER}}
Following the construction process, EMER contains a collection of 1,303 videos, totaling 390,900 frames. 
In EMER, we provide three distinct emotional signal subsets, namely the facial expression subset, the eye movement subset, and the emotion-related eye fixation subset, each of which is carefully designed to capture different aspects of emotions. Examples of the metadata in EMER are presented in Fig.~\ref{fig:example}, with additional samples provided in our \textit{supplementary material (Sec.III)}.


{\textbf{The facial expression subset}} comprises 1,303 videos and 390,900 frames. Each video has a duration of 1 to 2 minutes. {\textbf{The eye movement subset}} contains 1.91 million timestamp samples, offering a more comprehensive and richer set of information, such as time stamps, gaze point coordinates, gaze direction, pupil diameter, eye position, gaze time, and eye movement event type (sweep and gaze). 
{\textbf{The eye fixation subset}} is a massive collection of 7.50GB in size that contains 390,900 frames, each depicting a heat map of the video content, location, and trajectory of the participant's attention when emotional events occurred. 

Moreover, each emotional data in EMER possesses dual labels for FER and ER, respectively,  including 3-class FER and ER labels, 7-class FER and ER labels, 2-dimensional continuous emotion ratings, and facial expression intensity. 
Fig.\ref{fig:distri} reports the distributions of the rich ER and FER labels in our EMER. We also provide all \textit{manual} and \textit{automatic} annotations for each data in EMER. 

{\lkj In summary, EMER offers larger scale (1,303 samples from 121 participants), richer multi-view annotations (ER/FER, valence/arousal, and expression intensity), and non-invasive acquisition of synchronized facial and eye behaviors, making it a valuable benchmark for advancing research on FER and ER.}

\section{Proposed EMERT method}
Leveraging the EMER dataset, we design the Eye-behavior-aided MER Transformer (\textbf{EMERT}), incorporates Modality-Adversarial Feature Decoupling (MAFD) and an Emotion-sensitive Multi-task Transformer (EMT) to learn modality-complementary affective features, effectively bridging the emotional gap between facial expressions and eye behaviors for robust ER.   
As illustrated  in Fig.\ref{fig: model}, EMERT first employs a Multimodal Feature Extraction (MFE) module  to extract unimodal features. Then,  the MAFD applies a gradient reversal layer (GRL) with adversarial loss to decouple \textbf{emotion-generic features}, which are invariant to the emotion gap across various modalities, from \textbf{emotion-unique features}, which preserve their emotion discrepancies  across modalities.   Finally, the EMT further uses the  emotion-generic features as query to guide the fusion of emotion-unique features, enabling robust, emotion-sensitive representation learning for enhanced ER performance.  

\underline{\emph{MFE:}}
Given multimodal data as input, we employ a pre-trained Resnet \cite{ResNet} to extract the expression features $H_v\in \mathbb{R}^{T_v\times S}$, and use an LSTM \cite{LSTM1} to extract the eye movement features $H_e\in \mathbb{R}^{T_e\times S}$ and eye fixation features $H_g\in \mathbb{R}^{T_g\times S}$, respectively. $T_v$, $T_e$ and $T_g$ are the lengths of these three feature sequences, and $S$ is the dimension of feature vector.

\underline{\emph{MAFD:}}
With the multimodal features, we employ the MAFD module to separate the emotion-generic features $F_C$, invariant to the emotion gap across modalities, from the emotion-unique features $F_P$ retaining modality-specific emotional cues. 
To achieve this, we first apply an emotion-generic feature extractor (\textit{i.e.}, $MLP_c$) and three emotion-unique feature extractors (\textit{i.e.},  $MLP_{v}$, $MLP_{e}$, $MLP_{g}$), each implemented as a two-layer MLP. 
To ensure modality emotion invariance in $F_C$, we introduce a gradient reversal layer (GRL) after the $MLP_c$, and attach a modality classifier $D$ that distinguishes feature origins. 

By adversarially training $D$ and $MLP_c$, we encourage the $F_C$ to be indistinguishable across modalities,  while $F_P$ retains discriminative modality-specific cues.  The process can be formulated as, 
\begin{equation}
    \min_{\theta _D}\max_{\theta _{MLP_c}}\mathcal{L}_{adv}=
    -\frac{1}{N_b}\sum_{j=0}^{N_b}{f_j/e_j}\cdot \log F_D\left( F_C/F_P;\theta _D \right),
\end{equation}
where $N_b$ is the number of training samples,  $f_j/e_j$ represents the FER or ER labels, and $\theta_{MLP_c}$ and $\theta_D$ are the parameters of the $MLP_c$ and $D$, respectively.   
This adversarial setup ensures effective decoupling of emotion-generic and emotion-unique features, helping bridge the modality-induced emotion gap in ER.    
 
\underline{\emph{EMT:} }
Existing multimodal Transformer methods often uses queries from a single modality, which can overemphasize modality-specific cues and amplify the emotion gap between facial expressions and eye behaviors, leading to suboptimal fusion.   
To address this, EMT first adopts the decoupled emotion-generic features $F_C$ as query $q$, and the emotion-unique features $F_P$ as key $k$ and value $v$,  yielding more modality-complementary affective features. 
We follow the typical formulation of the Transformer structure as $Trans(\cdot)$ as:

\vspace{-0.5em}
\begin{equation}
    X_{fu}=Trans\left( q=F_C ,k/v=F_P \right). 
\end{equation}
With the modality-complementary affective features $X_{fu}$, we apply two multi-task prediction heads, namely the FER head and ER head, to predict the FER result $\hat{f_j}$ and the ER result $\hat{e_j}$, respectively.  
Each prediction head possesses a similar structure, comprising a 2-layer MLP.
Formally, the objectives for the two prediction heads are written as: 
\begin{equation}
	{{L}_{e}}=
 \left\{
\begin{array}{lr}   
    CE(\hat{e_j},{e_j}), & \text{for ~classification}
    \\
    Huber(\hat{e_j},{e_j}), & \text{for~regression}\\ 
    \end{array}
\right.
\end{equation}
\begin{equation}
	{{L}_{f}}=
 \left\{
\begin{array}{lr} 
    CE(\hat{f_j},{f_j}), & \text{for ~classification}
    \\
    Huber(\hat{f_j},{f_j}), & \text{for~regression}\\ 
    \end{array}
\right.
\end{equation}
For discrete emotion classification, we introduce the multi-class cross-entropy loss $CE()$; for continuous emotion regression, we employ the huber loss $Huber()$.

\underline{\emph{Overall Objective:}} The total objective function $\mathcal{L}$ of EMERT is the the summation  of the above-mentioned three learning objectives, which can be writen as: $\mathcal{L}=\alpha \mathcal{L}_{adv}+\beta (\mathcal{L}_{e}+\mathcal{L}_{f})$. 
Empirically, $\alpha=0.3 $ and $\beta=0.1$ are hyper-parameters to balance the multi-task learning process.
 
\vspace{-1em}
\section{EXPERIMENTS}

\subsection{\textbf{Experimental Setup}}

\subsubsection{\textbf{Evaluation Protocols}} 
To evaluate methods on the EMER dataset, we conducted both classification and regression protocols for both ER and FER tasks.
\textit{For the classification tasks}, consistent with the previous research, we chose three widely-used classification validation metrics, namely unweighted average recall (UAR), weighted average recall (WAR), and F-score (F1), to estimate our model. 
Larger values are preferred for all of these indicators.
\textit{For the regression tasks}, we also chose three widely-used regression validation metrics, like in other papers, namely Mean Absolute Error (MAE), Mean Squared Error (MSE), and Root Mean Squared  Error (RMSE). All of these regression metrics are as small as possible.

In addition, following existing evaluation protocols \cite{jiang20}, we employed a 5-fold cross-validation approach for these benchmarks on EMER. We utilized 1,043 data from the EMER dataset for training, and the remaining 260 data for testing.

\subsubsection{\textbf{Implementation Details}} 
In this paper, we used the PyTorch framework to implement all models on our EMER dataset. We set the batch size of all models to 16. All models were trained on an NVIDIA GeForce RTX 3090 with an initial learning rate as 0.0001. Cosine decay was used to decrease the learning rate during training. 
In the training and testing phases, each multimodal data in EMER is initially sampled with 8 frames evenly extracted from the facial expression video, 32 frames from the eye movement sequence, and 32 frames from the emotion-related eye fixation map. These frames are then input into the proposed EMERT model as well as other benchmark models to acquire emotion representations for predicting the final results.

\begin{figure*}[b]
  \centering
  \vspace{-1em}
  \includegraphics[width=0.95\linewidth]{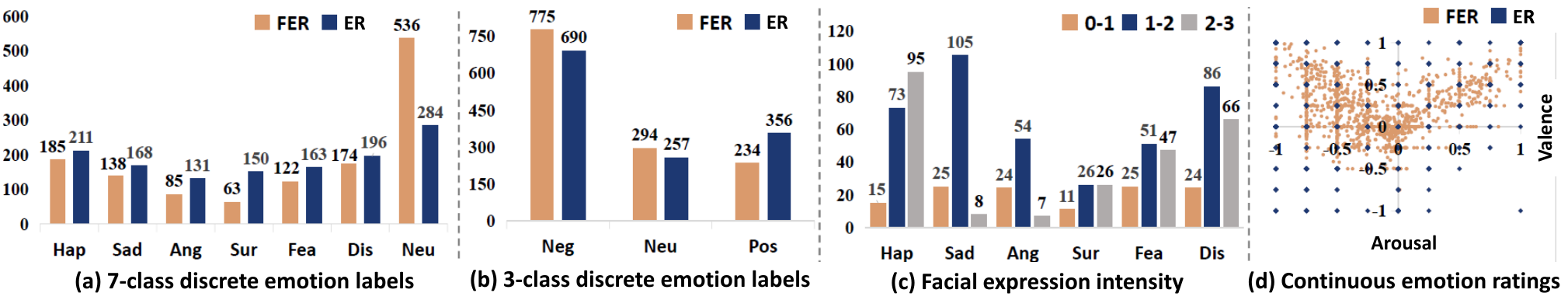}
  \vspace{-1em}
  \caption{The distribution differences between ER and FER labels in EMER. Hap, Sad, Ang, Sur, Fea, Dis, Neu, Neg, Pos are the abbreviations of the labels.}
  \label{fig:distri}
  \vspace{-1em}
\end{figure*}

\begin{table*}[!htb]
\vspace{-2em}
	\small
	\centering
	\caption{Comparison results of 3-class/7-class ER and FER on EMER, respectively.}
    \vspace{-1em}
	\label{tab:m_e3r}
	\scalebox{0.75}{
		\begin{tabular}{cccc|ccc|ccc|ccc}
			  \hline
              \multirow{3}*{Models}  & 
			  \multicolumn{6}{c|}{3-class}& 
			  \multicolumn{6}{c}{7-class} \\
              \cline{2-7} \cline{8-13}
			  & 
			  \multicolumn{3}{c|}{ER} & \multicolumn{3}{c|}{FER}& 
			  \multicolumn{3}{c|}{ER} & \multicolumn{3}{c}{FER}\\
			\cline{2-4} \cline{5-7} \cline{8-10} \cline{11-13}
			& WAR ($\uparrow$)  & UAR ($\uparrow$) & F1 ($\uparrow$)& WAR ($\uparrow$)  & UAR ($\uparrow$) & F1 ($\uparrow$)& WAR ($\uparrow$)  & UAR ($\uparrow$) & F1 ($\uparrow$)& WAR ($\uparrow$)  & UAR ($\uparrow$) & F1 ($\uparrow$) \\
			\hline
            TMC \cite{Han2022TrustedMC}&54.83	& 37.52  & 45.87&54.56	&39.84 &  46.96 & 31.28	 &27.20 &  25.20&49.64	 & 31.11   &42.66 \\
            TAILOR \cite{Zhang2022TailorVM}& 52.98  & 34.70 & 47.58& 63.91  & 43.61 & 53.12 & 33.39 & 24.85 & 28.40& 48.9  &30.55 &37.12\\
			MulT \cite{MulT}&  57.33 & 48.03 & 53.53&  64.94 & 52.18 & 61.48& 31.96 & 27.45 & 28.40& 48.83  & 32.82 & 42.78\\
			LMF \cite{LMF}   &  58.78 & \underline{49.36} & 54.80& 66.90 & \underline{56.55} & 63.64& 33.07  & 27.12&27.81& 49.34  & 32.05 & 42.53\\
		    ResNet\_LSTM  \cite{LSTM1}   & 57.78  &  49.06 & 55.08& \underline{67.24}  &  56.50 &  63.52& 31.18 & 27.64 & 27.48& 47.83  & 32.79  & 41.38\\
			ResNet\_Transformer \cite{ResNet} &  56.75 & 49.63 &  \underline{55.70}&  61.01 & 46.56 & 56.23& 29.21 & 23.49 & 25.98& 47.33  & 31.10 & 40.43\\
			C3D\_Transformer  \cite{Transformer}  & 52.51  &42.44 & 53.12& 59.67 &45.88 &55.01& 31.66  & 20.49&26.85& 42.50  &21.54 & 29.33\\
                Self\_MM \cite{Self_MM}   & 54.08  & 42.89 & 47.48& 61.51  & 42.34 & 50.69& 32.47 & 27.55 & 28.19& 49.58  & 32.05 & 42.77\\
			MMIM \cite{MMIM}   & 55.94  &46.93  & 53.13&68.05   & 55.99 & \underline{63.57}& 31.71 & 27.81 & 28.43& 48.61  & 31.77 & 42.50\\
                TMT \cite{TMT}   &  58.72  & 50.75  & 55.02  &   67.14 & 55.93  & 63.69 &  33.65  & 27.68  & 30.26  & 50.28   & 32.21  & 42.35 \\
                MMA-DFER \cite{MMA-DFER}   &  52.92    &  33.33   &   37.35  &  59.43   &  33.31  &  44.55 &  30.65  &  27.73  & \underline{30.42}  &  48.54  & \underline{32.81}  &  42.68\\
                NORM-TR \cite{NORM-TR}   &  \underline{59.13}  & 49.28  &  53.04 &  66.92  &  56.36 & 62.68 &  \underline{33.72}  & \underline{28.09}  &   28.81&  \underline{50.80}  & 32.63  & \underline{43.32}\\
            
\rowcolor{green!25}
			\textbf{EMERT} & \textbf{59.28}    & \textbf{52.62} & \textbf{55.71}& \textbf{68.10}    & \textbf{56.91} & \textbf{63.73}& \textbf{33.92}    & \textbf{28.17} & \textbf{30.38}& \textbf{51.18}    & \textbf{33.04} & \textbf{43.33}\\
			\hline
        \end{tabular}}
\end{table*}

\begin{table*}[!htb]
\vspace{-1em}
	\small
	\centering
	\caption{Comparison results of valence and arousal regression for ER and FER, respectively.}
    \vspace{-1em}
	\label{tab:m_va1r}
	\scalebox{0.75}{
		\begin{tabular}{cccc:ccc|ccc:ccc}
   \hline
			\multirow{3}*{Models} &\multicolumn{6}{c|}{ER} & \multicolumn{6}{c}{FER}\\
                 \cline{2-7}
                \cline{8-13}
			   & \multicolumn{3}{c:}{Arousal} &
                 \multicolumn{3}{c|}{Valence}&\multicolumn{3}{c:}{Arousal} &
                 \multicolumn{3}{c}{Valence}\\
			\cline{2-4}
                \cline{5-7}
                \cline{8-10}
                \cline{11-13}
			& MAE ($\downarrow$)&MSE ($\downarrow$)&RMSE ($\downarrow$)&MAE ($\downarrow$)&MSE ($\downarrow$)&RMSE ($\downarrow$) & MAE ($\downarrow$)&MSE ($\downarrow$)&RMSE ($\downarrow$)&MAE ($\downarrow$)&MSE ($\downarrow$)&RMSE ($\downarrow$)\\
			\hline
            TMC \cite{Han2022TrustedMC}&0.369	 &0.264 &0.468  &0.438&0.304&0.540&0.226	 &0.079  &0.276&0.375&0.218&0.460 \\
            TAILOR \cite{Zhang2022TailorVM}& 0.373 & 0.223 & 0.465 & 0.514 & 0.390 & 0.619& 0.246 & 0.089 &  0.294& 0.291 & 0.231 & 0.474\\
			MulT \cite{MulT}& 0.399 & 0.263 & 0.503 & 0.481 & 0.344 & 0.573& 0.236 & 0.085 & 0.287 & 0.338 & 0.180 &0.417\\
			LMF \cite{LMF}   &0.368 & \underline{0.219} & \underline{0.457} &0.440  &0.283  &0.526&0.228 & 0.080 & 0.278  & 0.286 &0.128  &0.359\\
		    ResNet\_LSTM \cite{LSTM1}   & 0.383 &  0.241& 0.482  & 0.454 & 0.303 & 0.543&0.229 & 0.082 & 0.281  & 0.303 & 0.138 & 0.366\\
			ResNet\_Transformer \cite{ResNet} & 0.388 & 0.241 & 0.481 & 0.450 & 0.293 & 0.537& 0.234 & 0.086 &0.288 & 0.299 & 0.136 & 0.362\\
			C3D\_Transformer  \cite{Transformer}  & 0.376 & 0.224 &0.465 &0.497 &0.333  &0.571&0.239  &0.087  & 0.288& 0.342 &0.179  &0.416\\
                Self\_MM \cite{Self_MM}   & 0.374 & 0.228 & 0.470 &  0.444 & 0.302 & 0.542& 0.244 & 0.092 & 0.297 & 0.351 & 0.189 &0.426\\
			MMIM \cite{MMIM}   &0.379  &0.230 & 0.471 &0.443 &0.295 &0.535&0.228  & 0.084 &0.285  &0.290  &0.131  &\underline{0.356}\\
            TMT \cite{TMT}   &  0.376  &  0.227 &  0.465 &  0.456  & 0.313  &  0.551  &  0.231  & 0.084  & 0.284  &  0.301  & 0.141  & 0.371 \\
                 MMA-DFER \cite{MMA-DFER}   &  0.366  & 0.235  & 0.493  &  \underline{0.440}  & 0.291  & 0.529   &  0.226  & 0.084  & \underline{0.273}  &  0.292  & \underline{0.128}  & 0.364 \\
                NORM-TR \cite{NORM-TR}   &  \underline{0.367}  &  0.221 & 0.461  &  0.437  & \underline{0.283}  &  \underline{0.521}  &  \underline{0.224}  & \underline{0.079}  &  0.278 &  \underline{0.289}  & 0.129  & 0.360 \\
\rowcolor{green!25}
 \textbf{EMERT} & \textbf{0.365} & \textbf{0.217}  &\textbf{0.456}  & \textbf{0.433} & \textbf{0.279} &\textbf{0.519}& \textbf{0.223} & \textbf{0.078}  &\textbf{0.266}  & \textbf{0.286} & \textbf{0.127} &\textbf{0.351}\\

			\hline
    \end{tabular}}
    \vspace{-1.5em}
\end{table*}

\vspace{-1em}
\subsection{\textbf{Benchmarking ER \& FER on EMER Dataset}}

We conducted extensive benchmarks for ER and FER, respectively, 
on the EMER dataset. 
For ER, we evaluate classification (3-class and 7-class) and valence/arousal regression. For FER, we further include intensity regression in addition to classification (3-class and 7-class) and valence/arousal regression.
In each task, we compared our EMERT with various cutting-edge methods including TMC \cite{Han2022TrustedMC}, TAILOR \cite{Zhang2022TailorVM}, MulT \cite{MulT}, LMF \cite{LMF}, ResNet\_LSTM \cite{ResNet}, Self\_MM \cite{Self_MM}, MMA-DFER \cite{MMA-DFER}, and so on. The results show that EMERT consistently outperforms these methods across all evaluation settings, demonstrating its effectiveness in handling eye behaviors information and mitigating the emotion gap.

\subsubsection{\textbf{3-class ER Classification}} 
Following \cite{vasileios_skaramagkas_2022_7794625}, we benchmarked EMERT against state-of-the-art methods on 3-class ER, as shown in Table~\ref{tab:m_e3r}. EMERT achieved the highest accuracy of 59.28\% WAR, 52.62\% UAR, and 55.71\% F1 on ER, improving over Self\_MM\cite{Self_MM} by 5.2\%, 9.73\%, and 8.23\%, respectively. 

\subsubsection{\textbf{7-class ER Classification}} 
In this setting, we compared our EMERT method with other typical and popular methods for 7-class ER. Table~\ref{tab:m_e3r} reports the comparison results of these benchmarks on EMER. EMERT achieved the highest accuracy of 59.28\% WAR, 52.62\% UAR, and 55.71\% F1 on ER, improving over Self\_MM\cite{Self_MM} by 5.2\%, 9.73\%, and 8.23\%, respectively. It indicates that EMERT can obtain more modality-complementary affective features for robust ER performance.

\subsubsection{\textbf{ER Valence/Arousal Regression}} 
Table~\ref{tab:m_va1r} reports the regression results on valence/arousal for ER. Unlike classification, regression tasks favor lower metric values. Regarding the performance on valence and arousal MSE and MAE, we observed that our EMERT obtained more precise regression results with smaller errors.
Moreover, EMERT achieves 0.9\% and 1.1\% lower MAE, and 1.1\% and 2.3\% lower MSE than Self\_MM\cite{Self_MM}, showing its fine-grained emotion perception. 

\subsubsection{\textbf{3-class FER Classification}} 
We compared EMERT with state-of-the-art methods for 3-class FER on  EMER dataset, and the results are shown in the Table ~\ref{tab:m_e3r}. EMERT achieved the significant gains over Self\_MM\cite{Self_MM}: 6.59\% WAR, 14.57\% UAR, and 13.04\% F1, demonstrating the effectiveness of integrating eye behavior cues for FER.

\subsubsection{\textbf{7-class FER Classification}} 
The performance results of the cutting-edge methods and our proposed EMERT method for 7-class FER are shown in Table~\ref{tab:m_e3r}. EMERT achieves the highest WAR (51.18\%), UAR (33.04\%), and F1 (43.33\%), indicating a stronger ability to capture facial expressive behaviors from multi-source signals.

\subsubsection{\textbf{FER Valence/Arousal Regression}} 
Table ~\ref{tab:m_va1r} provides the results of our EMERT and other cutting-edge methods for FER valence/arousal regression on EMER. Compared with TMT\cite{TMT}, EMERT achieves up to 0.8\% and 1.5\% improvement on MAE, 0.6\% and 1.4\% on MSE, and 1.8\% and 2.0\% on RMSE, confirming its robustness under continuous affective dimensions.

\subsubsection{\textbf{FER Intensity Regression}} 
To the best of our knowledge, we are the first to propose benchmarks for FER intensity regression. Table~\ref{tab:m_in} shows that EMERT outperforms all Transformer-based methods, achieved the gains over MMA-DFER\cite{MMA-DFER}: 2.6\% MAE, 2.2\% MSE, and 0.2\% RMSE, demonstrating the effectiveness of integrating eye behavior cues for FER.

\begin{table}[h]
\vspace{-0.5em}
	\small
	\centering
	\caption{Comparison of facial expression intensity regression for FER. }
    \vspace{-1em}
	\label{tab:m_in}
	\scalebox{0.75}{
		\begin{tabular}{ccccc}
			\toprule 
			\multirow{2}*{Models}  & 
			  \multicolumn{3}{c}{Metric} \\
			\cmidrule(r){2-4} 
			& MAE ($\downarrow$) & MSE ($\downarrow$)& RMSE ($\downarrow$)\\
			\midrule
			
			ResNet\_Transformer \cite{Transformer}& 0.701  & 0.707 & 0.829\\
			
			MulT \cite{MulT} & 0.804  & 1.035  & 0.999  \\
            TMT \cite{TMT} &  0.668  & 0.690   &  0.818  \\
            MMA-DFER \cite{MMA-DFER} & 0.686  &   0.695 &  \underline{0.811}  \\
            NORM-TR \cite{NORM-TR} &  \underline{0.666}  &  \underline{0.685}  &  0.823  \\
            \rowcolor{green!25}
			\textbf{EMERT} & \textbf{0.660}    & \textbf{0.673} & \textbf{0.809}\\
			\bottomrule
        \end{tabular}}
        \vspace{-1em}
\end{table}

\vspace{-1em}
\subsection{\textbf{Deepen Understanding of the EMER Dataset}}

\subsubsection{\textbf{Analysis of  ER labels and FER labels}}

To illustrate the gap between ER and FER, Fig.\ref{fig:distri} displays the distribution discrepancy between ER labels and FER labels in our EMER dataset. As shown, the "Neutral" category is significantly more dominant in FER labels, while high-intensity emotions such as "Surprise" are notably underrepresented. This suggests that FER tends to capture surface-level facial cues, whereas ER reflects individuals’ subjective emotional experiences. Such differences indicate a clear discrepancy in emotional perception between the two tasks.



\begin{figure*}[b]
  \centering
  \vspace{-1em}
  \includegraphics[width=0.95\linewidth]{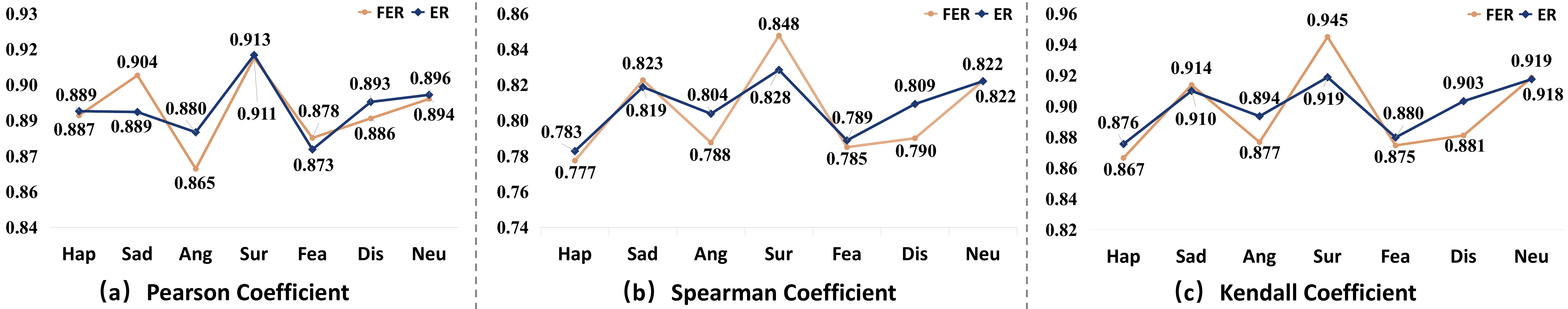}
  \vspace{-1em}
  \caption{The correlation analysis of eye behavior data for each category in ER and FER by three different correlation coefficients, \textit{i.e.}, (a) Pearson's coefficient, (b) Spearman's coefficient, and (c) Kendall's coefficient.
  The orange line illustrates the correlation of eye movement data for 7-class FER annotations, while the blue line signifies the correlation of eye movement data for 7-class ER annotations. In both cases, the closer the value is to 1, the stronger the correlation. }
  \label{fig: Correlations2}
\end{figure*}

\subsubsection{\textbf{Analysis of Benefits from Eye Behaviors for ER/FER}}
To explore the effect of eye behaviors in EMER for both ER and FER tasks, we used our proposed EMERT to perform separate emotion and faical expression classification with different modality settings on EMER, as shown in Table~\ref{tab:m_va}. The results show that using all modalities, including eye movement sequences, eye fixation maps and facial expression videos, yielded the most superior performance, underscoring that eye behaviors effectively enhance both ER and FER performance.
In addition, we observed that adding the eye movement sequences led to the highest improvement (see the underline results in the table), verifying that the eye behaviors are the effective complement of facial expressions for robust ER. 
We also investigate the correlations between eye behaviors and facial expressions for ER and FER tasks by using three distinct correlation coefficients: Pearson Coefficient, Spearman Coefficient, and Kendall Coefficient\cite{obilor2018test}.

\begin{table}[htbp]
\vspace{-1em}
	\small
	\centering
	\caption{Ablation study on 7-class ER and FER tasks. F, E, G denote facial expressions, eye movements, and fixation maps, respectively.}
    \vspace{-1em}
	\label{tab:m_va}
 \scalebox{0.75}{
		\begin{tabular}{cccccccccc}
			\toprule 
			\multicolumn{3}{c}{Modality} & 
			  \multicolumn{3}{c}{ER task}  & 
			  \multicolumn{3}{c}{FER task} \\
			\cmidrule(r){1-3}  \cmidrule(r){4-6}  \cmidrule(r){7-9}
			F &E & G & WAR ($\uparrow$) & UAR ($\uparrow$) & F1 ($\uparrow$) & WAR ($\uparrow$) & UAR ($\uparrow$) & F1 ($\uparrow$)\\
			\midrule
			\checkmark  & & & 30.21 & 26.40 & 29.32 &46.95   &  27.21 & 34.98\\
			  & \checkmark & & 31.51 & 26.12 & 27.61 &40.80  &18.99 & 24.92\\
			  & & \checkmark & 28.20 & 27.20& 24.10 &41.98   & 22.74 & 28.72\\
			\checkmark  & \checkmark & & \underline{32.89} & 27.03 & \underline{29.89} &   \underline{49.35}  & \underline{32.25} & \underline{41.98}\\
		   \checkmark  &  & \checkmark &32.81&27.63 & 29.85& 47.29 & 27.75  & 35.09\\
      
			 & \checkmark &  \checkmark &31.88 & \underline{27.64}& 29.01&43.32 & 24.71  & 31.22 \\
\rowcolor{green!25}
		 \checkmark &  \checkmark &  \checkmark & \textbf{33.92}    & \textbf{28.17} & \textbf{30.38} & \textbf{51.18}    & \textbf{33.04} & \textbf{43.33}  \\
			\bottomrule
	\end{tabular}}
    \vspace{-1em}
\end{table}

As depicted in Fig.\ref{fig: Correlations2}, the majority of correlation coefficients associated with the ER task exhibit higher values than the FER task. This observation indicates that eye movement data serves as a valuable complement to ER.
Meanwhile, the correlation coefficients pertaining to facial expressions also demonstrate that eye movement data effectively contributes to the understanding of FER. By combining the information, we can achieve a more comprehensive analysis of both emotions and facial expressions. This complementarity not only deepens our understanding of the `\textit{\textbf{emotion gap}}' but also holds promise for enhancing the performance of emotion analysis systems in various domains.

\subsubsection{\textbf{Effects of Different Annotation Methods}}
To assess the efficacy of our ALA  annotation methods for FER, we employed Cronbach's Alpha \cite{zadeh2018multimodal} to evaluate the consistency of different annotation approaches, as presented in Table~\ref{tab:model_annotation}. 
The results reveal that model auto-annotation exhibits the lowest reliability, primarily due to dataset bias.
The manual expert annotation, commonly used in many current datasets \cite{zadeh2018multimodal}, is susceptible to subjective individual differences, such as identity and profession, resulting in an average low Cronbach's Alpha of 0.799.
The combination of model auto-annotation and expert annotation can enhance the quality and consistency of emotion labels (as seen in the third row of Table \ref{tab:model_annotation}).
Ultimately, our ALA approach achieves the most consistent and reliable FER labels, significantly surpassing other methods, owing to the incorporation of annotation reliability assessment.

\begin{table}[!ht]
\vspace{-1em}
    \small
	\caption{Annotation consistency evaluation on different annotation approaches. The best results are in bold.}
    \vspace{-1em}
    \centering
    \label{tab:model_annotation}
    \scalebox{0.8}{
    \begin{tabular}{c|c|c}
    \hline
        \normalsize{Methods} & \normalsize{Label Type} & \normalsize{Cronbach’s Alpha}  \\ 
        
        \hline
          & \normalsize{Discrete emotion category}  &  \normalsize{0.731} \\ 
       \normalsize{Model auto-annotation} & \normalsize{Valence rating}  &  \normalsize{0.789} \\ 
        ~ & \normalsize{Arousal rating} &  \normalsize{0.679} \\ 
        \hline
         & \normalsize{Discrete emotion category}& \normalsize{0.784}  \\ 
         \normalsize{Expert annotation}& \normalsize{Valence rating}  &  \normalsize{0.829} \\ 
        ~ & \normalsize{Arousal rating} &  \normalsize{0.786} \\
        \hline
         \multirow{3}{*}{\makecell{\normalsize{Expert annotation +}\vspace{3pt} \\ \normalsize{Model auto-annotation}}}& \normalsize{Discrete emotion category} & \normalsize{\underline{0.852}}  \\ 
         ~ & \normalsize{Valence rating}  &  \normalsize{\underline{0.863}} \\ 
        ~ & \normalsize{Arousal rating} &  \normalsize{\underline{0.847}} \\ 
        \hline
        \rowcolor{green!25}
          & \normalsize{Discrete emotion category} & \normalsize{\textbf{0.978}}  \\ 
          \rowcolor{green!25}
        \normalsize{Proposed ALA} & \normalsize{Valence rating}  & \normalsize{\textbf{0.927}}  \\ 
        \rowcolor{green!25}
        ~ & \normalsize{Arousal rating} & \normalsize{\textbf{0.982}}  \\ 
        \hline
    \end{tabular}}
\end{table}

\subsubsection{{\lkj \textbf{Effects of Different Eye Movement features}}}
{\lkj To further investigate the contribution of different Eye Movement information, we conducted an ablation study on EMER by progressively isolating individual features. Specifically, we evaluated models using only gaze point, gaze time, and pupil diameter, and compared them with our complete eye movement modeling design.
As shown in Table~\ref{tab:de}, the results reveal that different features contribute unequally to ER and FER. For ER, gaze point and pupil diameter provide more emotion-related information, while gaze time yields relatively weaker performance. For FER, pupil diameter achieves the most significant improvements, followed by gaze time, with gaze point being the weakest. Importantly, when integrating all three types of features, our complete design consistently outperforms single-feature settings on both ER and FER tasks.
These findings demonstrate that eye movement information not only carries meaningful affective cues but also provides complementary signals when different features are combined. This validates the necessity of modeling multiple eye movement features jointly, rather than relying on a single aspect, to enhance the robustness and accuracy of multimodal emotion recognition.}

\begin{table}[htbp]
	\small
	\centering
    \vspace{-1em}
	\caption{{\lkj Performance comparison of different eye movement features on the 7-class ER and FER tasks. The best results are in bold.}}
    \vspace{-1em}
	\label{tab:de}
  \scalebox{0.78}{
		\begin{tabular}{c|ccc|ccc}
            \hline
            \addlinespace[0.5ex]
			 \multirow{2}{*}{Type} & \multicolumn{3}{c|}{ER}& \multicolumn{3}{c}{FER}\\ 
           \cline{2-4}  \cline{5-7} 
           \addlinespace[0.5ex]
         & WAR ($\uparrow$) & UAR ($\uparrow$) & F1 ($\uparrow$)& WAR ($\uparrow$) & UAR ($\uparrow$) & F1 ($\uparrow$)\\
       \addlinespace[0.5ex]
         	\hline
        Gaze point  & 31.96 & \underline{27.17} & \underline{28.47} & 46.18  & 30.64  & 36.53 \\
        Gaze time    & 31.13  & 27.05 & 25.21 & 47.57  & 30.48  &  38.06 \\ 
        Pupil diameter   &  \underline{32.38}  & 26.85 & 27.42  & \underline{50.46} & \underline{31.27}  & \underline{41.34} \\
        \rowcolor{green!25}
           \textbf{Ours} &\textbf{33.92}&\textbf{28.17}& \textbf{30.38}& \textbf{51.18}    & \textbf{33.04} & \textbf{43.33}  \\
           
	
			\bottomrule
	\end{tabular}}
\end{table}

\subsection{\textbf{Deepen Understanding of the EMERT Method}}

We conduct extensive ablation studies to further validate our method, with feature distribution visualizations across tasks provided in the \textit{supplementary material (Sec.\~III)}. 

\subsubsection{\textbf{Analysis of Attention Gap between ER and FER}}
\begin{figure}[htbp]
\vspace{-1em}
  \centering
  \includegraphics[width=0.95\linewidth]{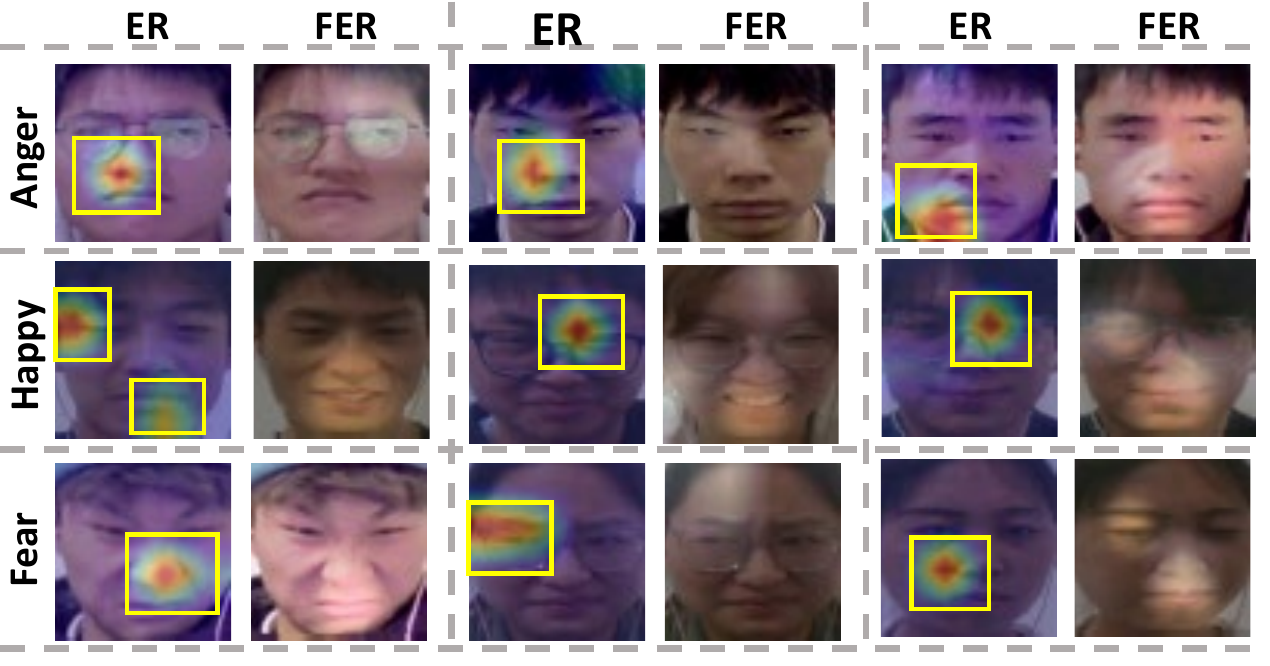}
  \vspace{-1em}
  \caption{Attention maps from the ER head and FER head in EMERT. For ER, the model pays more attention to detailed
areas such as the corners of the eyes, mouth, and nose, while for FER, the model pays more attention to the global areas, such as the entire eyes and mouth.}
  \label{fig: visiualization}
  \vspace{-1em}
\end{figure}

In addition to the label differences, we can further carry out deeper interpretation with our EMER dataset. In particular, by training an EMERT on our dataset, we can visualize the attention maps obtained by the FER and ER heads in EMERT, respectively, as illustrated in Fig.\ref{fig: visiualization}. According to the figure, it can be seen that the ER head pays more attention to detailed areas such as the corners of the eyes, mouth, and nose, while the FER head pays more attention to the global areas, such as the entire eyes and mouth.

\subsubsection{\textbf{Analysis of Whether FER and ER Reinforces Each Other}}
To investigate the potential benefits of ER and FER to each other, Table~\ref{tab:m_m} compared their performance in single-task and multi-task settings on the 3-class ER and FER tasks, respectively. 
Specifically, we first conducted the experiments on the FER task with and without the ER head integrated into our proposed EMERT model (see Fig. \ref{fig: model_structure}), respectively. Following this, we proceeded to evaluate the ER task with and without the FER head.
Our results revealed that the inclusion of the FER head in the ER task significantly enhances performance, resulting in a 5.94\% increase in the UAR metric. Additionally, the integration of the ER head also improves FER results by 0.37\% in WAR.
These experimental results highlight the mutually reinforcing effects of ER and FER, indicating that our proposed EMER and the corresponding EMERT method help to understand the emotion gap between emotions and facial expressions, ultimately improving state-of-the-art performance.

\begin{table}[htbp]
\vspace{-1em}
	\small
	\centering
	\caption{Inter-augmentation of FER and ER  in EMERT. }
    \vspace{-1em}
	\label{tab:m_m}
  \scalebox{0.8}{
		\begin{tabular}{c|cc|ccc}
			\hline
		\multirow{2}*{Type}  & \multicolumn{2}{c|}{Task head}  & \multicolumn{3}{c}{Metric}\\ 
           \cline{2-3}  \cline{4-6}
        & FER & ER& WAR ($\uparrow$) & UAR ($\uparrow$) & F1 ($\uparrow$)\\
         	\hline
			Single-task FER & \textcolor{black}{\checkmark} & \textcolor{black}{X} & 67.73  & \hl{\textbf{58.31}}  & 63.18 \\
			Multi-task FER &\textcolor{black}{\checkmark} & \textcolor{black}{\checkmark} & \hl{\textbf{68.10}}  & 56.91 & \hl{\textbf{63.73}} \\\hline
			Single-task ER& \textcolor{black}{X} &\textcolor{black}{\checkmark} &  56.01  & 46.68 & 50.85 \\
			Multi-task ER& \textcolor{black}{\checkmark} & \textcolor{black}{\checkmark} & \hl{\textbf{59.28}}  & \hl{\textbf{52.62}} & \hl{\textbf{55.71}} \\
			\bottomrule
	\end{tabular}}
    \vspace{-1em}
\end{table}

\subsubsection{\textbf{Evaluation on the Another SIMS Dataset}}



To validate EMERT's effectiveness, we followed \cite{Self_MM} and evaluated our EMERT on the SIMS dataset using Acc-2/3/5. SIMS \cite{SIMS} provides video, audio, and text with multimodal and unimodal annotations. We utilize 1,824 samples for training and 457 samples for testing. As shown in Table~\ref{tab:Results_Compare_SIMS}, EMERT outperforms the other Transformer-based method, \textit{i.e.}, MulT\cite{MulT}, with a relative increase of 1.86\% (Acc-2), 3.34\% (Acc-3), and 23.04\% (Acc-5), highlighting its effectiveness and generalization.

\begin{table}[h]
	\small
    \vspace{-1em}
	\caption{Comparison results of multimodal emotion recognition on SIMS. The best results are in bold.}
 \vspace{-1em}
 \label{tab:Results_Compare_SIMS}
	\centering
	\resizebox{0.8\linewidth}{!}
	{
    	\begin{tabular}{cccc}
            \toprule
    		Models & Acc-2 ($\uparrow$)&  Acc-3 ($\uparrow$)& Acc-5 ($\uparrow$)\\
     		\midrule

            MulT \cite{MulT} & 78.84  & \underline{67.13}  & 38.24 \\
    		TFN \cite{TFN} &  \textbf{82.06} & 66.16 & 39.74 \\  
    		MFN \cite{MFN} & 78.26  & 65.79 & 41.19 \\ 
    		Self-MM \cite{Self_MM} & 78.99  & 66.52 & \underline{44.20} \\ 
            \rowcolor{green!25}
    		\textbf{EMERT} & \underline{80.31} &  \textbf{69.37}  & \textbf{47.05} \\ 	
            \bottomrule
    	\end{tabular}
	}
    \vspace{-1em}
\end{table}

\subsubsection{\textbf{Effects of Different Modules in EMERT}}

To assess each module’s impact, we conduct ablation studies on the 7-class ER and FER tasks using the EMER dataset (Table~\ref{tab:modular}).  
The baseline method, a Transformer-based multimodal fusion approach \cite{Transformer}, directly combines facial expression features from the pre-trained ResNet \cite{ResNet} with eye movement and eye fixation features from the pre-trained LSTM \cite{LSTM1}.
{\lkj Firstly, with the addition of the MAFD module, we observe consistent improvements compared to the baseline. Specifically, there is a relative increase of 2.31\% in ER and 0.50\% in FER for WAR. This demonstrates that modality-adversarial decoupling effectively reduces non-emotional interference and helps the model learn clearer modality-invariant emotion representations.  
Secondly, the EMT module further enhances performance with a relative increase of 8.19\% in ER and 1.76\% in FER for F1, indicating its capability to exploit complementary supervision from ER and FER labels. By jointly modeling both tasks, EMT alleviates task-specific bias and strengthens the cross-task generalization of the learned features.  
Ultimately, our full EMERT model, combining MAFD and EMT, achieves the best performance across all metrics, confirming that these modules are complementary and together contribute to both ER and FER tasks.}

\begin{table}[htbp]
	\small
	\centering
    \vspace{-1em}
	\caption{Module ablation study on the 7-class MER and FER, respectively. The best results are in bold.}
    \vspace{-1em}
	\label{tab:modular}
  \scalebox{0.69}{
		\begin{tabular}{ccc|ccc|ccc}
            \hline
            \addlinespace[0.5ex]
			 \multicolumn{3}{c|}{Module} & \multicolumn{3}{c|}{ER}& \multicolumn{3}{c}{FER}\\ 
           \cline{1-3}  \cline{4-6} \cline{7-9}
           \addlinespace[0.5ex]
       Baseline & MAFD & EMT & WAR ($\uparrow$) & UAR ($\uparrow$) & F1 ($\uparrow$)& WAR ($\uparrow$) & UAR ($\uparrow$) & F1 ($\uparrow$)\\
       \addlinespace[0.5ex]
         	\hline
         \checkmark & &  & 31.18 &27.64&27.48&47.83 & 32.79 & 41.38  \\
        \checkmark & \checkmark &  &31.90 & 27.85& 29.70&  48.07 & 32.85 & \underline{42.80}  \\ 
        \checkmark & & \checkmark  &\underline{32.93}  &\underline{28.00}&\underline{29.73}&\underline{48.61} & \underline{32.91} & 42.11 \\
        \rowcolor{green!25}
           \checkmark & \checkmark & \checkmark &\textbf{33.92}&\textbf{28.17}& \textbf{30.38}& \textbf{51.18}    & \textbf{33.04} & \textbf{43.33}  \\
           
	
			\bottomrule
	\end{tabular}}
    \vspace{-1em}
\end{table}

\subsubsection{\textbf{Robustness of EMERT against Noisy Data}}
To evaluate robustness, we injected Gaussian noise with varying variances into each modality for the 3-class emotion recognition task. As shown in Table~\ref{tab:denose}. Notably, existing methods such as MMIM~\cite{MMIM} and Self-MM~\cite{Self_MM} exhibit a significant performance drop (over 5\%) under noise. In contrast, EMERT shows only a 2.29\% decrease in WAR when the variance is set to 0.01, with a slight increase in F1. These results demonstrate that EMERT is more resilient to modality-specific noise compared to prior approaches.

\begin{table}[!ht]
\vspace{-1em}
    \small
	\caption{Robustness comparison of different methods for 3-class ER on EMER. The best results are in bold.}
    \vspace{-1em}
    \centering
    \label{tab:denose}
    \scalebox{0.9}{
    \begin{tabular}{cccc}
    \hline
    \addlinespace[0.5ex]
        \multirow{3}*{Methods} & \multicolumn{3}{c}{Variance (Gaussian) setting }\\
        \cmidrule(){2-4} 
        &0.01(WAR/F1) & 0.05(WAR/F1) & 0.1(WAR/F1)  \\ 
         \midrule
        MMIM \cite{MMIM} & \underline{46.23/45.02} & \underline{45.13/40.05} & \underline{42.82/36.01}  \\ 
        Self-MM \cite{Self_MM} & 42.98/42.24 & 42.27/42.16 & 37.20/40.45  \\ 
        \rowcolor{green!25}
        EMERT & \textbf{56.99/57.01} & \textbf{55.42/57.30} & \textbf{56.33/56.23}  \\ 
         \midrule
    \end{tabular}}
\end{table}

\subsubsection{{\lkj \textbf{Effects of Different hyperparameter in EMERT}}}

{\lkj To investigate the sensitivity of EMERT to hyperparameter settings, we conduct experiments on the 7-class ER and FER tasks with varying values of the adversarial loss weights $\alpha$ and $\beta$. As shown in Table~\ref{tab:hyper}, the performance exhibits notable variation under different configurations.  
When $\alpha=0.3$ and $\beta=0.1$, EMERT achieves the best results, with WAR/UAR/F1 reaching 33.92/28.17/30.38 for ER and 51.18/33.04/43.33 for FER, respectively. This suggests that a moderate adversarial loss weight provides an effective balance: it is strong enough to guide modality-adversarial decoupling, but not so dominant as to destabilize training. In contrast, excessively small weights (e.g., $\alpha=0.1$ or $\beta=0.01$) reduce the effectiveness of the decoupling mechanism, while overly large weights (e.g., $\alpha=0.5$ or $\beta=1$) can overwhelm the supervised learning objective, leading to degraded performance.  
These findings confirm that EMERT is relatively robust to hyperparameter variations within a reasonable range, and that carefully tuning $\alpha$ and $\beta$ further enhances the balance between adversarial feature decoupling and multi-task learning, thereby improving overall performance.}

\begin{table}[htbp]
	\small
	\centering
    \vspace{-1em}
	\caption{{\lkj Hyperparametric analysis experiment on the 7-class ER and FER, respectively. The best results are in bold.}}
    \vspace{-1em}
	\label{tab:hyper}
  \scalebox{0.8}{
		\begin{tabular}{cc|ccc|ccc}
            \hline
            \addlinespace[0.5ex]
			 \multirow{2}{*}{$\alpha$}  & \multirow{2}{*}{$\beta$}  & \multicolumn{3}{c|}{ER}& \multicolumn{3}{c}{FER}\\ 
           \cline{3-5}  \cline{6-8} 
           \addlinespace[0.5ex]
        & & WAR ($\uparrow$) & UAR ($\uparrow$) & F1 ($\uparrow$)& WAR ($\uparrow$) & UAR ($\uparrow$) & F1 ($\uparrow$)\\
       \addlinespace[0.5ex]
         	\hline
         ~ & 0.01  & 30.19  & 26.31 & 26.51 & 45.31 &  29.83 &  38.13  \\
       0.1 & 0.1    & 31.69  & \underline{27.44} & 27.28 & 46.64  &  32.26 & 40.43  \\ 
        ~ & 1   & 31.2  & 26.57 & 26.36 & 48.28 & 32.13  & 40.47 \\
        \midrule
         ~ & 0.01  &  28.67 & 24.39 & 21.57 & 44.66 & 26.38  &  34.31  \\
        \rowcolor{green!25}
       0.3 & 0.1    & \textbf{33.92}  & \textbf{28.17} & \textbf{30.38} &  \textbf{51.18} & \textbf{33.04}  & \textbf{43.33}  \\ 
        ~ & 1   & \underline{33.09}  & 27.36 & \underline{29.73} & \underline{51.15} & \underline{32.8}  &  \underline{40.71}\\
        \midrule
         ~ & 0.01  & 25.14  & 19.59 & 16.68 & 41.79 &  19.58 &  26.75  \\
       0.5 & 0.1    & 26.19  & 20.63 & 18.56 & 43.71 &  23.57 & 30.42  \\ 
        ~ & 1   & 29.55   & 23.77  & 21.79  &  47.84 & 28.93   &  36.93 \\
	
			\bottomrule
	\end{tabular}}
\end{table}

\vspace{-1em}
\section{CONCLUSIONS AND FUTURE WORK}
In this paper, we explored the gap between Facial Expression Recognition (FER) and Emotion Recognition (ER) by introducing eye behaviors as a crucial emotional cue. To support this, we constructed the Eye-behavior-aided Multimodal Emotion Recognition (EMER) dataset, which contains 1,303 spontaneous emotional samples from 121 participants. Notably, EMER provides multi-view emotion labels for both ER and FER, enabling a comprehensive analysis to elucidate the gap between them. 
Building upon EMER, we propose the Eye-behavior-aided MER Transformer (EMERT), a simple yet effective model that integrates modality-adversarial feature decoupling and multitask learning to effectively fuse eye behaviors and facial expressions. Extensive experiments across seven benchmark protocols demonstrate that EMERT significantly outperforms state-of-the-art multimodal methods, highlighting the importance of modeling eye behaviors for robust and complementary emotion understanding.
This paper provides a complete framework—from dataset to model—for advancing multimodal emotion recognition. Our findings offer insights into the gap between FER and ER, and emphasize the value of eye behaviors in enhancing emotional perception. 
{\lkj In the future, we plan to extend EMER with more diverse and ecologically valid scenarios and release it to encourage further research in emotion-related tasks. Moreover, the proposed EMERT framework shows strong potential for real-world applications such as human–computer interaction, mental health monitoring, and emotionally intelligent virtual agents. We also aim to integrate large multimodal models to further investigate the relationship between FER and ER. These directions will not only enrich the academic value of this work but also enhance its practical impact.}

\vspace{-1em}
\bibliographystyle{IEEEtran}
\bibliography{IEEEabrv,software}

\vfill

\end{document}